\def\year{2020}\relax
\documentclass[letterpaper]{article} 
\usepackage{aaai20}  
\usepackage{times}  
\usepackage{helvet} 
\usepackage{courier}  
\usepackage[hyphens]{url}  
\usepackage{graphicx} 
\usepackage{graphicx}  
\title{Automatic Text-based Personality Recognition on Monologues and Multiparty Dialogues Using Attentive Networks and Contextual Embeddings}
\author{\Large \textbf{Hang Jiang, \textsuperscript{\rm 1} Xianzhe Zhang,\textsuperscript{\rm 2} Jinho D. Choi\textsuperscript{\rm 3}}\\ 
\textsuperscript{\rm 1}Symbolic Systems Program, Stanford University, CA 94305\\
\textsuperscript{\rm 2}Department of Electrical Engineering, Stanford University, CA 94305\\
\textsuperscript{\rm 3}Department of Computer Science, Emory University, Atlanta, GA 30322\\
hjian42@stanford.edu,
xianzhez@stanford.edu, 
jinho.choi@emory.edu
}

\begin{document}
\maketitle
\begin{abstract}
Previous works related to automatic personality recognition focus on using traditional classification models with linguistic features. However, attentive neural networks with contextual embeddings, which have achieved huge success in text classification, are rarely explored for this task. In this project, we have two major contributions. First, we create the first dialogue-based personality dataset, \verb|FriendsPersona| , by annotating 5 personality traits of speakers from Friends TV Show through crowdsourcing. Second, we present a novel approach to automatic personality recognition using pre-trained contextual embeddings (BERT and RoBERTa) and attentive neural networks. Our models largely improve the state-of-art results on the monologue Essays dataset by 2.49\%, and establish a solid benchmark on our \verb|FriendsPersona|. By comparing results in two datasets, we demonstrate the challenges of modeling personality in multi-party dialogue. 
\end{abstract}

\section{Introduction}
Automatic text-based personality recognition, as an important topic in computational psycho-linguistics, focuses on determining one's personality traits from text. The Big Five Hypothesis is usually used for measuring one's personality in five binary traits: \textit{agreeableness (AGR), conscientiousness (CON), extraversion (EXT), openness (OPN), neuroticism (NEU)}. Recently, \citeauthor{majumder2017deep} (2017) use Convolutional Neural Networks (CNN) with static word embeddings and outperform the previous feature-based systems \cite{mairesse2007using,tighe2016personality} on Essays dataset \cite{pennebaker1999linguistic}. However, previous works have neither explored dialogue data nor use attentive networks and contextual embeddings such as BERT \cite{devlin2018bert} and RoBERTa \cite{liu2019roberta} for the task.

To address these issues, we create the first dialogue dataset \verb|FriendsPersona| for automatic personality recognition with a novel and scalable dialogue extraction algorithm, \textit{MainSpeakerFinder}. Besides, we introduce both attentive networks and contextual embeddings (BERT and roBERTa) to the task. We not only outperform the previous models on the benchmark Essays dataset, but also achieve strong baseline results on our \verb|FriendsPersona| dataset.

\section{Dataset}

We focus on Essays and \verb|FriendsPersona| datasets. Essays Dataset \cite{pennebaker1999linguistic} is the benchmark dataset for text-based personality recognition with 2,468 self-report essays. Our new \verb|FriendsPersona|\footnote{\url{https://github.com/emorynlp/personality-detection}} dataset is developed upon the public Friends TV Show Dataset \cite{chen2016character} and contains 711 extracted conversations. Each essay or conversation in the two datasets is annotated by the binary Big Five personality traits. 

\subsubsection{MSF Extraction Algorithm}
To build our own dataset, we develop a novel dialogue extraction algorithm, \textit{MainSpeakerFinder} (MSF), to extract sub-scenes from full scenes and mark each sub-scene with a main speaker for three annotators to annotate. First of all, we slide a window of size 5 across the full dialogue to track the utterance count per speaker at each step. This allows us to obtain a smoothed utterance count curve per speaker. Then, we find peaks in each speaker's utterance count curve. At last, we extract conversations around peaks as sub-scenes, in which the speaker of the curve is always the main speaker. This extraction step is necessary for two reasons. First, it allows annotators to focus on a short dialogue text. Moreover, the algorithm reuses full scenes to generate many short sub-scenes, which is beneficial to building a comparatively large dataset for training. 

\subsubsection{Annotation Processing}
Due to limited funding, we annotated 711 sub-scenes from the first 4 seasons of the Friends TV Show on Amazon Mechanical Turk. Each sub-scene is annotated by 3 annotators for Big Five personality traits with -1, 0, and 1. We sum scores from 3 annotators and convert them to binary class with the median split. 

\subsubsection{Inter-Annotator Agreement}
In terms of inter-annotator agreement, we achieve an average pair-wise kappa of 54.92\% between 2 annotators and Fleiss' kappa of 20.54\% among 3 annotators across five personality traits. The low Fleiss' kappa is rather expected because text-based personality recognition is highly subjective so that annotators often judge different personality traits that are all acceptable for the same utterance. This may also be attributed to the limitation of our data; a higher agreement could be achieved if a multimodal dataset (e.g. text, image, audio) is provided.

\section{Experiments}
\subsection{Data Preparation}
To be consistent with previous works \cite{majumder2017deep}, Essays and \verb|FriendsPersona| datasets use accuracy and 10-fold cross validation with a constant seed for sampling. In \verb|FriendsPersona|, we replaced speaker names with marks like 'speaker0' and 'speaker1'. Both datasets have binary class labels for each personality trait.

\subsection{Experiment on Essays Dataset}
First, we experiment baseline models on Essays dataset with FastText embeddings. In Table \ref{tab:baseline-table}, Majority represents the percentage of the dominant class. LIWC (2016) represents the best LIWC-based model's performance \cite{tighe2016personality}. HCNN is Hierarchical CNN model \cite{majumder2017deep}. ABCNN and ABLSTM represents CNN and Bidirectional LSTM models with attention mechanism. HAN is Hierarchical Attention Network. Besides, we fine-tuned pre-trained base BERT and RoBERTa embeddings for the task. Overall, ABCNN achieves the best score on CON, whereas RoBERTa gets the best on the other four traits. We improve 2.49\% for 5 traits on average (AGR by 2.22\%, CON by 2.83\%, EXT by 2.53\%, OPN by 3.18\%, NEU by 1.69\%).


\begin{table}[]
\centering
\resizebox{.95\columnwidth}{!}{
\begin{tabular}{c|ccccc}
\textbf{Models} & \textbf{AGR}   & \textbf{CON}   & \textbf{EXT}   & \textbf{OPN}   & \textbf{NEU}   \\ 
\hline
Majority        & 53.08          & 50.81          & 51.74          & 51.54          & 50.04          \\
\hline
LIWC (2016)     & \textbf{57.50}    & 56.00             & 55.70             & 61.95             & 58.30          \\
HCNN (2017)     & 56.71             & \textbf{57.30}    & \textbf{58.09}    & \textbf{62.68}    & \textbf{59.38}          \\
\hline
ABCNN           & 57.82          & \textbf{60.13} & 58.75          & 63.65          & 58.51          \\
ABLSTM          & 58.85          & 59.55          & 59.32          & 63.99          & 59.56 \\
HAN             & 57.62          & 59.32          & 59.77          & 63.61          & 58.75          \\
BERT            & 58.10         & 57.69           & 59.12         & 61.17           & 59.20 \\
RoBERTa         & \textbf{59.72}         & 58.55           & \textbf{60.62}         & \textbf{65.86}           & \textbf{61.07} \\
\end{tabular}
}
\caption{The performance of models in accuracy on Essays.}\smallskip
\label{tab:baseline-table}
\end{table}


\subsection{Adaptation to FriendsPersona Dataset}
We also experiment these models on \verb|FriendsPersona|. We experiment with three ways of feeding dialogue text to the classifiers: 1. \textit{single (S)}: use only the concatenation of the single target speaker's utterances; 2. \textit{single + context (S+C)}: use \textit{S} + the concatenation of other speakers' utterances; 3. \textit{full (F)}: use the full dialogue text in the natural order. 

First, the models perform the best on \textit{S} out of 3 formats for 5 traits (Table \ref{tab:diag-table}). It makes sense because our models are originally designed to classify simple monologue text instead of multi-party dialogue text and \textit{S} converts dialogue to the target speaker's monologue. Second, BERT and RoBERTa together achieve the most best results (10 out of 15 cases). But BERT and RoBERTa do not beat other models for CON on both datasets. At last, HAN achieves 3 best results out of 15 cases on \verb|FriendsPersona|, better than its performance (0 case) on Essays dataset. This is because HAN encodes dialogue on both utterance and token levels, which allows HAN to attend the main speaker's utterances. In the future, we need a customized model to leverage dialogue information between speakers. 


\begin{table}[]
\centering
\resizebox{.95\columnwidth}{!}{
\begin{tabular}{c|ccccccc}
Trait        & Format & Majority & ABCNN & ABLSTM         & HAN            & BERT           & RoBERTa        \\ \hline
             & S      & 56.96    & 63.86 & 64.56          & 64.00          & 62.02          & \textbf{65.58} \\
\textbf{AGR} & S+C    & 56.96    & 59.64 & 60.76          & \textbf{61.60} & 59.77          & 57.77          \\
             & F      & 56.96    & 59.21 & 62.01          & 61.88          & 62.77          & \textbf{64.49} \\ \hline
             & S      & 53.59    & 56.40 & 57.38          & \textbf{58.66} & 55.21          & 56.78          \\
\textbf{CON} & S+C    & 53.59    & 54.71 & 57.53          & 57.53          & \textbf{57.77} & 55.92          \\
             & F      & 53.59    & 54.99 & 56.67          & \textbf{57.81} & 57.07          & 57.35          \\ \hline
             & S      & 56.12    & 59.78 & 59.50          & 60.35          & 61.77          & \textbf{64.21} \\
\textbf{EXT} & S+C    & 56.12    & 59.64 & \textbf{62.03} & 57.25          & 60.34          & 59.05          \\
             & F      & 56.12    & 58.93 & \textbf{61.60} & 58.37          & 63.62          & 60.05          \\ \hline
             & S      & 64.98    & 65.40 & 66.52          & 67.23          & 67.19          & \textbf{68.47} \\
\textbf{OPN} & S+C    & 64.98    & 66.95 & 66.10          & 66.67          & \textbf{67.61} & 66.90          \\
             & F      & 64.98    & 66.39 & 66.52          & 66.39          & 66.33          & \textbf{67.19} \\ \hline
             & S      & 53.31    & 56.54 & 57.52          & 58.23          & 59.06          & \textbf{60.06} \\
\textbf{NEU} & S+C    & 53.31    & 57.12 & 59.22          & 59.21          & \textbf{60.20} & 58.76          \\
             & F      & 53.31    & 57.82 & 58.66          & 58.36          & 56.49          & \textbf{59.33}
\end{tabular}
}
\caption{The performance on FriendsPersona with 3 formats.}\smallskip
\label{tab:diag-table}
\end{table}

\section{Conclusions}

In this paper, we have two major contributions. We create a new dialogue corpus \verb|FriendsPersona| for automatic personality recognition with a novel and scalable MSF dialogue extraction algorithm. Besides, we introduce both attentive neural networks and contextual embeddings to the task. We significantly outperform the state-of-art results on the monologue Essays dataset, and establish a solid benchmark on \verb|FriendsPersona|. In the future, we want to design a BERT-based attention network to model utterances in dialogue and improve the performance on our dataset. Besides, we plan to assign more annotators to improve annotation quality and to expand the corpus size.


\bibliographystyle{aaai}
\bibliography{refs}

\end{document}